\def\year{2020}\relax
\documentclass[letterpaper]{article} 
\usepackage{aaai20}  
\usepackage{times}  
\usepackage{helvet} 
\usepackage{courier}  
\usepackage[hyphens]{url}  
\usepackage{graphicx} 
\usepackage{graphicx}  

\usepackage{url}
\usepackage{verbatim}
\usepackage{multirow}
\usepackage{subfigure}
\usepackage{algorithm}
\usepackage{algorithmicx}
\usepackage{amsmath,amsfonts,amssymb,bbm,epsfig,bm}
\usepackage{algpseudocode}

\usepackage{soul}
\usepackage{xcolor}
\usepackage{color}
\newcommand{\fancyhl}[3][red]{\colorlet{#1#2}{#1!#2}\sethlcolor{#1#2}\hl{#3}}

\title{Recursive Graphical Neural Networks for Text Classification}
\author{Wei Li\textsuperscript{1}, Shuheng Li\textsuperscript{1}, Shuming Ma\textsuperscript{1} Yancheng He\textsuperscript{2}, Deli Chen\textsuperscript{1}, Xu Sun\textsuperscript{1,3} \\
\textsuperscript{1}MOE Key Lab of Computational Linguistics, School of EECS, Peking University \\
\textsuperscript{2}Platform \& Content Group, Tencent \\
\textsuperscript{3}Deep Learning Lab, Beijing Institute of Big Data Research, Peking University\\
\texttt{\{liweitj47, shuhengli, shumingma\}@pku.edu.cn}\\
\texttt{\{collinhe\}@tencent.com} \\
\texttt{\{chendeli, xusun\}@pku.edu.cn} \\}

\begin{document}

\maketitle

\begin{abstract}
  The complicated syntax structure of natural language is hard to be explicitly modeled by sequence-based models. 
  Graph is a natural structure to describe the complicated relation between tokens. The recent advance in Graph Neural Networks (GNN) provides a powerful tool to model graph structure data, but simple graph models such as Graph Convolutional Networks (GCN) suffer from over-smoothing problem, that is, when stacking multiple layers, all nodes will converge to the same value. In this paper, we propose a novel \textbf{Re}cursive \textbf{G}raphical \textbf{N}eural \textbf{N}etworks model (ReGNN) to represent text organized in the form of graph. In our proposed model, LSTM is used to dynamically decide which part of the aggregated neighbor information should be transmitted to upper layers thus alleviating the over-smoothing problem. Furthermore, to encourage the exchange between the local and global information, a global graph-level node is designed. We conduct experiments on both single and multiple label text classification tasks. Experiment results show that our ReGNN model surpasses the strong baselines significantly in most of the datasets and greatly alleviates the over-smoothing problem.
  
\end{abstract}

\section{Introduction}
Neural models are the dominant approach in many NLP tasks. It is an essential step to represent text with a dense vector for many NLP tasks, such as text classification \cite{liu2016recurrent} and summarization \cite{see2017get}. Traditional methods represent text with hand-crafted sparse lexical features, such as bag-of-words and n-grams \cite{wang2012baselines,silva2011symbolic}. Recently, deep learning models have been widely used to learn text representations, including convolutional neural networks (CNN) \cite{kim2014convolutional} and recurrent neural networks (RNN) such as LSTM \cite{hochreiter1997long}. As CNN and RNN prioritize locality and sequentiality \cite{battaglia2018relational}, these deep learning models tend to capture local consecutive information well. However, the dependency between words often exceeds the scope of local windows. Sequential information flow results in weaker ability in capturing long range dependencies, which leads to lower performance when encoding long sentences \cite{koehn2017six}. 

The ability to organize text with graph structure makes the model more powerful in representing the text.
\citeauthor{peng2018large} argue that constructing word graph with word co-occurrence information can effectively model the relation among words. However, GNN models such as GCN often suffer from the over-smoothing problem \cite{li2018deeper,zhou2018graph}, that is to say, when stacking multiple layers, all nodes will converge to very similar values. On the other hand, the shallow network structure limits the expression power significantly. The design of the gate mechanism in LSTM is powerful to control the information flow between the new input and the previous hidden states, which makes LSTM a possible solution to the over-smoothing problem. \citeauthor{song2018graph} propose to apply LSTM to encode the AMR graph, which provides a good view on how to apply LSTM on graphs.  

In this paper, we propose a recursive graphical neural networks model to encode the text graph constructed from the word co-occurrence information.  LSTM is used to filter the aggregated information calculated based on the attention on the neighbors and update the hidden states of the nodes. A graph-level node is further added to represent the whole graph and interact with the ordinary nodes. 

We conduct experiments on both single and multiple label text classification to verify the effectiveness of our proposed ReGNN model. Experiment results show that our model surpasses several strong baselines on most of the widely used benchmark datasets significantly and largely overcomes the problem of over-smoothing faced by many graph based models.

We conclude our contributions as follows:
\begin{itemize}
    \item We propose a recursive graphical neural networks model to represent text graph that dynamically decides how to aggregate and update the hidden state of the nodes, which largely overcomes the over-smoothing problem.
    \item We propose a way to organize the text sequence into a graph, which can be easily modeled with graph networks.
    \item We conduct extensive experiments on both single and multiple label text classification tasks. Experiment results show that our model surpasses most of the strong baselines significantly.
\end{itemize}

\section{Related Work}
LSTM has been successfully applied in many NLP tasks such as machine translation \cite{bahdanau2014neural}, summarization\cite{see2017get}, text classification \cite{liu2016recurrent} and so on. The gate mechanism in LSTM is powerful in modeling long sequences by controlling the information flow with dynamically calculated weights. Although under many circumstances LSTM can handle the information flow in text sequence, there exists long dependency among words that linear LSTM can not explicitly model but are suitable for graph models. Furthermore, using the last hidden state (known as last pooling) to represent the whole sentence is also problematic. The information exchange between the word nodes and the global node in our proposed model makes the word nodes aware of the global information and the global node can dynamically decide which nodes are more important.

Convoluational neural networks (CNN) \cite{kim2014convolutional} focus on modeling the local features by applying filters on local text windows. This model can capture the semantic information of word and phrase level. However, long dependencies are hard to be modeled by local filters even with pooling. Our proposed model considers the local features by treating the consecutive words in the sequence as neighbors in the graph. The aggregation of neighbor information then provides the ability to model the local features. Apart from local connections, our proposed model can also model longer dependency and global information that CNN fails to model.

Attention mechanism \cite{bahdanau2014neural} is a good way to model long dependency. Transformer \cite{vaswani2017attention} applies self-attention instead of recurrent connection to handle the dependency between words. However, in a text sequence, not all the words are semantically connected, which means the involvement of all the words in the attention can introduce much noise. In our proposed model, we use attention mechanism to aggregate the neighbor information in combination with the gate mechanism to filter the newly aggregated information. This design makes the model able to deal with various number of neighbors and dynamically decide which part of the information should be updated.

Graph structure is powerful for modeling complicated word connections. 
\citeauthor{peng2018large} propose a graph CNN model to first convert text to graph-of-words, which is then used as input to graph convolution operations in \citeauthor{niepert2016learning}.  \citeauthor{yao2018graph} regard both the documents and the words as nodes to construct a heterogeneous graph and uses  GCN \cite{kipf2016semi} to learn embeddings of words and documents. Our model also adopts the idea of a global node in the graph. The introduction of LSTM makes the updating of our model more powerful than GCN thus avoiding the over-smoothing problem. 

\section{Approach}

\begin{figure}[t]
    \centering
    \includegraphics[width=8cm]{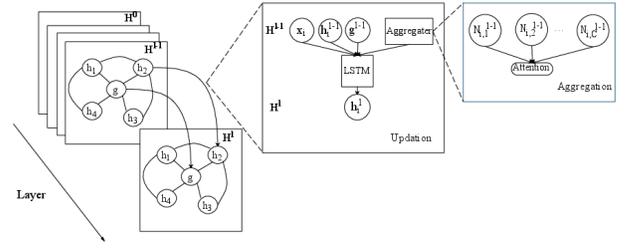}
    \caption{Illustration of the process of how the graph representation is calculated in the proposed Recursive Graphical Neural Networks, which mainly consists of \textbf{Aggregation} and \textbf{Updation} two parts.  On the right we give detailed descriptions for the \textbf{updation} part and the \textbf{aggregation} part. \textbf{Aggregation} part decides how to aggregate the information from the neighbors of a node. \textbf{Updation} part decides how to update the hidden state at the $l$-th layer based on the newly aggregated neighboring information. $H^l$ represents the hidden states of all nodes at  $l$-th layer.  $h_i$ is the hidden state of the word node $i$. $x_i$ is the embedding of the $i$-th node. $\mathcal{N}_{i,j}$ indicates the hidden state of the $j$-th neighbor of the node $i$. $g$ is the graph-level node that represents the whole graph.}
    \label{fig:model structure}
\end{figure}

In this section, we describe our proposed recursive graphical neural networks model. A brief illustration is shown in Figure \ref{fig:model structure} and Algorithm \ref{alg:graph_construction}. We first convert the text into a graph based on the word co-occurrence information. LSTM is used to decide how to update the hidden states of the nodes. Attention mechanism is used to aggregate the neighbor information of each node. To capture the global information and allow each node to interact with the global information, a graph-level node is added.

\subsection{Text Graph Construction}
\label{sec:text_graph}

In this part we describe how to convert text into a graph based on the word co-occurrence information. It should be noted that the construction of the text graph is not restricted to word co-occurrence information, other methods such as dependency parsing can also be applied. If the text consists of multiple sentences, we apply NLTK toolkit\footnote{\url{https://www.nltk.org/}} to segment the text into sentences and tokenize the sentence into words. If the sentence structure is not available, a sliding window is applied to extract the structure. Given a text span $s$ (sentence or text within a sliding window) with $m$ words $w_1, w_2, \cdots, w_m$, each word is a node in the sentence graph.  The point-wise mutual information (PMI) is used to decide whether the two words should be connected in the graph. 
\begin{equation}
    PMI(w_i, w_j) = log\frac{p(w_i, w_j)}{p(w_i) p(w_j)}
\end{equation}
 If $w_i$ and $w_j$ appear in the same sentence, the count of co-occurrence of $w_i$ and $w_j$ is added by one. We count the co-occurrence between words based on all the text in the corpus. If the PMI score between $w_i$ and $w_j$ is positive, $w_j$ is added to the neighbor candidates of $w_i$. Furthermore, to capture the local context of words, the direct neighbors $w_{i-1}$ and $w_{i+1}$ are also connected with the word $w_i$. The candidate neighbors with at most $n-2$ highest positive PMI scores and the two direct neighbors are taken as the real neighbors of $w_i$, which are connected with $w_i$ in the graph. 
 
 \begin{algorithm}[t]
\centering
\footnotesize
\begin{algorithmic}[1]
\Require
The input text $s$. PMI function to build text graph.
\State
Tokenize the text $s$ into $m$ words. 
\State 
Construct a text graph based on the PMI.
\State
\For{node $i$ in all the nodes in the graph}
    \State Look up the embedding of node $i$ and get $x_i$
    \State Initialize the hidden state $h_i$ of each node with $x_i$
\EndFor
\For{layer $l$ in $\left\lbrace 1,2,\cdots,L\right\rbrace$}
    \State initialize the graph-level node vector $g$ with Eqn. (\ref{eqn:pooling_1}-\ref{eqn:pooling_3})
    \For{node $i$ in all the nodes in the graph}
        \State Aggregate the neighbor information of node $i$ with Eqn. (\ref{eqn:agg_1}-\ref{eqn:agg_4}) and get $\mathcal{N}_i$
        \State Update the $h_i^l$ based on LSTM with Eqn. (\ref{eqn:update_1}-\ref{eqn:update_6})
    \EndFor
    \State Update $g^l$ with Eqn. (\ref{eqn:pooling_1}-\ref{eqn:pooling_3})
\EndFor
\State
\State Classify the text with $g^L$ and $h_1,\cdots,h_m$
\end{algorithmic}
\caption{Process of ReGNN for Text Classification.}
\label{alg:graph_construction}
\end{algorithm}

\subsection{Recursive Graphical Neural Networks}
In this part, we describe the structure of the Recursive Graphical Neural Networks model we propose. Given a text graph consisting of $m$ nodes, $w_1, w_2, \cdots, w_m$, we not only learn the representation of each node, but also learn the representation $g$ of the whole graph. 

\subsubsection{Updation}
\noindent We first get the embeddings of each node in the graph, which gives $\left \langle x_1, x_2, \cdots, x_m \right \rangle$. Rather than applying LSTM sequentially along the text sequence like traditional linear LSTM model, the recurrent structure is applied layer wise, which follows the way of gated graph neural networks \cite{li2015gated}. Formally, at the $l$-th layer, the hidden representation of the text graph is denoted as:
$$H^l = \left \langle h^l_1, h^l_2, \cdots, h^l_m, g^l\right \rangle$$
where $g$ is the representation of the graph-level node. $h$ is the hidden state of each node. For the initial state $H^0$, the hidden states of the $i$-th node is set to its embedding, $h^0_i=x_i$. The transition from $h^{l-1}_i$ to $h^l_i$ is calculated as follows:
\begin{align}
i^l_i &= \sigma (W_i [h^{l-1}_i; x_i; g^{l-1}; \mathcal{N}_i^{l-1}] + b_i)  \label{eqn:update_1} \\ 
f^l_i &= \sigma (W_f [h^{l-1}_i; x_i; g^{l-1}; \mathcal{N}_i^{l-1}] + b_f)  \label{eqn:update_2} \\ 
o^l_i &= \sigma (W_o [h^{l-1}_i; x_i; g^{l-1};  \mathcal{N}_i^{l-1}] + b_o) \label{eqn:update_3}  \\ 
u &= tanh (W_u [h^{l-1}_i; x_i; g^{l-1}; \mathcal{N}_i^{l-1}] + b_u) \label{eqn:update_4}  \\ 
c^l_i &= f^l_i \odot c^{l-1}_i + i^{l-1}_i \odot u \label{eqn:update_5} \\ 
h^l_i &= o^l_i \odot tanh(c^l_i) \label{eqn:update_6}
\end{align}
where $\mathcal{N}_i$ denotes the aggregated hidden representation of the neighbors of the node $w_i$. $[\ ;\ ;\ ]$ means the concatenation of vectors. At each layer, the usage of node embedding $x_i$ is to make use of the original meaning of the word, which serves as the role similar to residual connections. By using the representation of the graph-level node $g$, the network is able to model the interaction between each node and the whole graph, which makes it better able to focus on the important information. $i, f, o$ indicate the \textit{input}, \textit{forget} and \textit{output} gates respectively. 

\subsubsection{Aggregation}
\noindent Unlike S-LSTM \cite{zhang2018sentence} that only considers the direct left and right neighbors $w_{i-1}$ and $w_{i+1}$, our proposed model is able to take flexible number of neighbors when aggregating the neighbor information. We propose to use additive attention \cite{bahdanau2014neural} to aggregate the neighbor information. Assuming that $w_i$ has $C_i$ neighbors in the text graph, the hidden representation $\mathcal{N}_i$ of the neighbors of $w_i$ is calculated as follows:
\begin{eqnarray}
& \mathcal{P}_i = h_i + p_i \label{eqn:agg_1} \\ 
& \alpha_j = u(W_n[ \mathcal{P}_i; x_i; g; \mathcal{P}_j] + b_n) \label{eqn:agg_2} \\ 
& score_j = \frac{exp(\alpha_j)}{\sum_k^{C_i} exp(\alpha_k)} \label{eqn:agg_3} \\ 
& \mathcal{N}_i = \sum_k^{C_i} score_k h_{i_k} \label{eqn:agg_4}
\end{eqnarray}
where $p_i$ is the position embedding of the node $w_i$, $h_{i_k}$ is the hidden state of the $k$-th neighbor of node $w_i$. By applying positional encoding in the attention, the model can be aware of the position information of word $i$ and $j$. Note that the positional embedding is optional, if the graph is not constructed out of a text sequence, positional embedding is not needed. The combination of LSTM and attention based aggregation makes our model able to gather information from longer dependency as the layer number increases while determining which part of the information is helpful and should be passed to higher layers. 

\subsubsection{Graph-level Node}
\noindent The value of $g^l$ is computed based on the hidden states from the last layer $H^{l-1}$. To get an overview of $H^{l-1}$, $\bar{h}^{l-1}$ is first calculated by doing attention on $\left \langle h_1^{l-1}, \cdots, h_m^{l-1} \right \rangle$. 
\begin{eqnarray}
& \alpha_i = u(W_a h_i) \label{eqn:pooling_1} \\  
& score_i = \frac{exp(\alpha_i)}{\sum_j\exp(\alpha_j)} \label{eqn:pooling_2}  \\ 
& \bar{h} = \sum_j score_j h_{j} \label{eqn:pooling_3}
\end{eqnarray}
Then for each $h_i^{l-1}$, a gate $f_i^l$ is calculated to decide which information should be considered by the graph-level vector $g^l$. The candidate state $c_g^l$ is calculated based on the candidate graph state at last layer $c_g^{l-1}$ and the  candidate states of the nodes at last layer $c_i^{l-1}$.
\begin{align}
& \hat{f}_g^l = \sigma (W_g [g^{l-1}; \bar{h}^{l-1}] + b_g) \\
& \hat{f}_i^l = \sigma (W_f [g^{l-1}; h_i^{l-1}] + b_f) \\
& o^l = \sigma (W_o [g^{l-1}; \bar{h}^{l-1}] + b_o) \\
& f^l_0, \cdots, f^l_{m}, f_g^l = softmax(\hat{f}_0^l, \cdots, \hat{f}^l_{m},\hat{f}_g^l) \\
& c_g^l = f^l_g \odot c_g^{l-1} + \sum_i f_i^l \odot c_i^{l-1} \\
& g^l = o^l \odot tanh(c^l_g)
\end{align}

\subsubsection{Comparison with Linear LSTM}
\noindent Linear LSTM (including the bidirectional version) is widely used in NLP applications to model a sequence. Linear LSTM takes the input one by one and outputs the hidden state of each input token iteratively. The theoretical time complexity is $O(N)$, where $N$ is the length of the input sequence. Different from the linear LSTM, suppose the memory of the computer is big enough, our model can take in the whole graph at the same time and the theoretical \textbf{time complexity} is $O(1)$, which means constant time. The exact time is dependent on the hyper-parameter of the layer number, which is a constant number. There is no need to wait for the calculation of the hidden states of the previous token like linear LSTM does. The calculation of the hidden state only needs the hidden states of the graph from the last layer $H^{l-1}$, while the hidden states of the nodes at the same layer are calculated concurrently, which is similar to the stacked CNN model. Furthermore, the popular pooling methods applied in linear LSTM (last pooling, max pooling, mean pooling) suffer from bias, while the graph-level node in ReGNN can interact with other ordinary nodes at each layer.

\subsubsection{Comparison with Transformer and GAT}
The core idea behind the Transformer model is using attention to build the connection between the current token and its context. The similar part between ReGNN and Transformer is that both of the models apply attention mechanism to learn the context information. However, the practical application scenarios between the two models are very different. ReGNN is used for graph-structure data, while Transformer is used for linear-structure data. Only when enforcing the graph to be fully connected, Transformer can be adopted, which may actually bring in much noise. In the graph scenario, the GAT (Graph Attention network) model actually resembles the idea of Transformer. However, both GAT and Transformer do not have specific design to prevent the over-smoothing problem and there is no graph-level node involved.

\subsection{Text Classifier}
For the single label task, a simple linear layer is applied to predict the label $y$ based on the graph-level vector $g^L$ ($L$ is the number of layers) calculated as follows:
\begin{equation}
    y = argmax(W_{out} g^L)
\end{equation}

For the multiple label task, we adopt the recurrent neural network (RNN) decoder with attention mechanism that can generate labels of various lengths \cite{SGM}. 
We adopt this framework to predict multiple labels because of its flexibility for generating various number of labels without tuning the threshold hyper-parameter.

Given the initial state $t_0$ which is the vector for the whole graph $g^L$ and the vectors of each node $\left \langle h_1^L, h_2^L, \cdots, h_m^L \right \rangle$, the decoder is bound to generate a sequence of labels $y_1, y_2, \cdots, y_k$. At each decoding step, a context vector $c_i$ is calculated by doing attention on the vectors of the nodes,
\begin{align}
    y_i &= argmax(W_{out} t_i)   \\
    t_i &= LSTM(t_{i-1}, E y_{i-1}) \\
    c_i &= \sum\alpha_j\times h^L_j \\
    \alpha_j &= \frac{exp(\delta(t_i, h_j^L)}{\sum exp(\delta(t_i, h^L_k))}
\end{align}
where $\delta$ is the attention function. A $\left \langle START\right \rangle$ and an $\left \langle END\right \rangle$ token are added to the front and the back of the label sequence. The generation process finishes when the model encounters the $\left \langle END\right \rangle$ token.

For both of the classification tasks, a standard cross entropy loss is minimized by an optimizer to train the model.

\section{Experiment}
We test the performance of our ReGNN model on two kinds of text classification tasks, namely, single label classification and multiple label classification.

\subsection{Datasets}
For the single label classification task, we run experiments on five widely used benchmark corpora including R8, R52, Ohsumed, Stanford Sentiment Treebank (SST) and Movie Review (MR). The statistics are shown in Table \ref{tab:stats single}.
\begin{itemize}
    \item The Ohsumed corpus\footnote{\url{http://disi.unitn.it/moschitti/corpora.htm}} is from the MEDLINE database. 
    Each document in the set has one or more associated categories from 23 disease categories. The 7,400 documents belonging to only one category are used. 
    \item R8 and R52\footnote{\url{https://www.cs.umb.edu/˜smimarog/textmining/datasets/}} (all-terms version) are two subsets of the Reuters21578 dataset. R8 has 8 categories and R52 has 52 categories, which are all associated with a single topic.
     \item MR is a movie review dataset for binary sentiment classification, in which each review only contains one sentence \cite{pang2005seeing}.\footnote{\url{http://www.cs.cornell.edu/people/pabo/movie-review-data/}} 
    We use the train/test split in \citeauthor{tang2015pte}.\footnote{\url{https://github.com/mnqu/PTE/tree/master/data/mr}}
    \item  Stanford Sentiment Treebank (SST) is an extension of MR but with train/dev/test splits provided and fine-grained labels (very positive, positive, neutral, negative, very negative), re-labeled by \citeauthor{socher2013recursive}.
   
\end{itemize}

\begin{table*}[t]
    \centering
    \begin{tabular}{c|ccccccc}\hline
        Dataset & \# Docs  & \# Train & \# Dev & \# Test & \# Words & \# Classes & Average Length \\ \hline
        Ohsumed & 7,400 & 3,022 & 335 & 4,043 & 14,157 & 23 & 135.82 \\ 
        R8 & 7,674 & 4,936 & 548 & 2,189 & 7,688 & 8 & 65.72 \\
        R52 & 9,100 & 5,878 & 653 & 2,568 & 8,892 & 52 & 69.82 \\
        SST & 11,855 & 8,544 & 1,101 & 2,210 & 4,683 & 5 & 19.17\\
        MR & 10,661 & 8,529 & 1,066 & 1,066 & 18,764 & 2 & 20.39 \\ \hline
    \end{tabular}
    \caption{Statistics of single label text classification. For sst and MR, we use the predefined dev set. For other datasets, we randomly split 10\% of the training set as dev set.}
    \label{tab:stats single}
\end{table*}

For the multiple label classification task, we run experiments on two widely used benchmark corpora including 
Reuters21578 and Ohsumed (full version). The statistics of the datasets are shown in Table \ref{tab:stats multi}.
\begin{itemize}

    \item Reuters21578:\footnote{\url{http://kdd.ics.uci.edu/databases/Reuters21578/Reuters21578.html}} This dataset consists of documents that appeared on the Reuters Newswire in 1987. We use the Lewis modified data split for fair comparison.
    \item Ohsumed: we use all the abstracts and their corresponding labels instead of only using documents with one label.
   
\end{itemize}
\begin{table*}[t]
    \centering
    \begin{tabular}{c|cccccccc}\hline
        Dataset & \# Docs  & \# Train & \# Dev & \# Test & \# Words & \# Classes & \# Ave Label & \# Ave Word \\ \hline
        Reuters & 10,377 & 6,362 & 706 & 3,309 & 10,294 & 119 & 1.26 & 126.59 \\
        Ohsumed & 13,929 & 5,658 & 628 & 7,643 & 22,974 & 23 & 1.66 & 179.48 \\
     \hline
    \end{tabular}
    \caption{Statistics of multiple label text classification. 
    We randomly split 10\% of the training set as dev set.}
    \label{tab:stats multi}
\end{table*}

\begin{table*}[t]
    \centering
    \begin{tabular}{c|ccccc} \hline
     & R8 & R52 & Ohsumed & SST & MR \\ \hline
     CNN  &  97.42 $\pm$ 0.41 & 94.11 $\pm$ 0.23 & 67.90 $\pm$ 0.38 & 42.30 $\pm$ 0.41 & 77.49 $\pm$ 0.30 \\
     RCNN    & 97.27 $\pm$ 0.33 & 93.59 $\pm$ 0.63 & 65.53 $\pm$ 1.07 & 43.23 $\pm$ 0.44 & 78.26 $\pm$ 0.56 \\
     BiLSTM  &  97.14 $\pm$ 0.30 & 92.89 $\pm$ 0.26 & 55.42 $\pm$ 0.84 & 42.63 $\pm$ 0.56 & 77.33 $\pm$ 0.85 \\
     2-BiLSTM  & 96.90 $\pm$ 0.60 & 92.61 $\pm$ 0.81 & 51.83 $\pm$ 1.76 & 42.40 $\pm$ 0.64 & 77.29 $\pm$ 0.50 \\ 
     Transformer & 94.99 $\pm$ 0.43 & 86.72 $\pm$ 1.01 & 36.45 $\pm$ 1.33 & 40.63 $\pm$ 1.23 & 74.85 $\pm$ 1.05 \\
     PV-DBOW & 85.87 $\pm$ 0.10 & 78.29 $\pm$ 0.11 & 46.65 $\pm$ 0.19 & 38.12 $\pm$ 0.33 & 61.09 $\pm$ 0.10 \\
     PV-DM & 52.07 $\pm$ 0.04 & 44.92 $\pm$ 0.05 & 29.50 $\pm$ 0.07 & 35.93 $\pm$ 0.68 & 59.47 $\pm$ 0.38 \\
     PTE & 96.69 $\pm$ 0.13 & 90.71 $\pm$ 0.14 & 53.58 $\pm$ 0.29 & 37.57 $\pm$ 0.60 & 70.23 $\pm$ 0.36 \\
     FastText & 96.13 $\pm$ 0.21 & 92.81 $\pm$ 0.09 & 57.70 $\pm$ 0.49 & 36.08 $\pm$ 0.84 & 75.14 $\pm$ 0.20 \\
     SWEM &  95.32 $\pm$ 0.26 & 92.94 $\pm$ 0.24 & 63.12 $\pm$ 0.55 & 42.08 $\pm$ 0.94 & 76.65 $\pm$ 0.63 \\
     LEAM & 93.31 $\pm$ 0.24 & 91.84 $\pm$ 0.23 & 58.58 $\pm$ 0.79 & 42.93 $\pm$ 0.79 & 76.95 $\pm$ 0.45 \\
     Graph-CNN-C & 96.99 $\pm$ 0.12 & 92.75 $\pm$ 0.22 & 63.86 $\pm$ 0.53 & 35.25 $\pm$ 0.52 & 77.22 $\pm$ 0.27 \\
     Graph-CNN-S & 96.80 $\pm$ 0.20 & 92.74 $\pm$ 0.24 & 62.82 $\pm$ 0.37 & 35.67 $\pm$ 0.76 & 76.99 $\pm$ 0.14 \\
     Graph-CNN-F & 96.89 $\pm$ 0.06 & 93.20 $\pm$ 0.04 & 63.04 $\pm$ 0.77 & 25.99 $\pm$ 2.65 & 76.74 $\pm$ 0.21 \\
     Text GCN & 97.07 $\pm$ 0.10 & 93.56 $\pm$ 0.18 & \textbf{68.36 $\pm$ 0.56} & 40.63 $\pm$ 0.13 & 76.74 $\pm$ 0.20   \\
     HAN & - & - & 67.75 $\pm$ 0.84 & - & - \\
     S-LSTM & 97.57 $\pm$ 0.23 & 94.25 $\pm$ 0.37 & 64.47 $\pm$ 0.99 & 42.46 $\pm$ 0.78 & 77.06 $\pm$ 0.39 \\
     ReGNN & \textbf{97.93* $\pm$ 0.31} & \textbf{95.17* $\pm$ 0.17} & 67.93 $\pm$ 0.33 & \textbf{43.93* $\pm$ 0.41} & \textbf{78.71* $ \pm$ 0.56} \\ 
     \hline
    \end{tabular}
    \caption{Experiment results on single label text classification. We run all models 10 times and report mean $\pm$ standard deviation.  $*$ indicates significance based on student t-test (p $<$ 0.055). Because the punctuation in R8 and R52 (all terms version) is cleaned in the original resource, hierarchical structures can not be applied in R8 and R52. Because texts in SST and MR consist of one single sentence, hierarchical structure can not be applied.}
    \label{tab:single result}
\end{table*}
\begin{table*}[t]
    \centering
    \begin{tabular}{c|ccc|ccc} \hline
     \multirow{2}{*}{}&\multicolumn{3}{c|}{Reuters21578} & \multicolumn{3}{c}{Ohsumed} \\ \cline{2-7}
     & Precision & Recall & F1 & Precision & Recall & F1 \\ \hline
     CNN   & 82.27 & 71.71 & 76.63 & 35.28 & 31.65 & 33.37 \\
     RCNN  & 72.57 & 61.71 & 66.70 & 22.60 & 27.13 & 24.65 \\
     Graph-CNN & 73.72 & 66.69 & 70.03 & 38.64 & 27.62 & 32.32 \\
     BiLSTM & 80.71 & 72.52 & 76.40 & 47.68 & 42.91 & 45.17 \\
     2-BiLSTM & 81.69 & 73.94 & 77.62 & 54.40 & 50.48 & 52.36 \\ 
     Transformer & 50.34 & 45.15 & 47.60 & 40.55 & 36.28 & 38.30 \\
     HAN & 83.65 & 77.88 & 80.66 & 55.07 & \textbf{53.34} & 54.19 \\
     S-LSTM & 83.26 & 75.01 & 78.92 & 54.26 & 50.07 & 52.08 \\
     GraphSAGE & 80.32 & 71.75 & 75.79 & 36.01 & 27.49 & 31.18 \\
     ReGNN & \textbf{84.69} & \textbf{79.48} & \textbf{82.01}* & \textbf{56.03} & 53.00 & \textbf{54.47}* \\ 
     \hline
    \end{tabular}
    \caption{Experiment results on multiple label text classification. \textit{Chebyshev} filter is used for Graph-CNN. We run the models with top 2 F1 scores 10 times, $*$ indicates significance based on student t-test (p $<$ 0.05)}
    \label{tab:multi result}
\end{table*}

\subsection{Baselines}
\begin{itemize}
    \item \textbf{CNN}: We employ Convolutional Neural Networks  \cite{kim2014convolutional} with the suggested window sizes in the paper.
    \item \textbf{Bi-LSTM} \cite{liu2016recurrent}: A bi-directional LSTM, commonly used in text classification. 
    \item \textbf{PV}: A Paragraph Vector model proposed by \citeauthor{le2014distributed}. The order of words is considered in DW setting while ignored in DBOW setting. We use Logistic Regression as the classifier.
    \item \textbf{PTE}: Predictive Text Embedding \cite{tang2015pte}, which averages the word embeddings trained based on the heterogeneous text network containing words, documents and labels as nodes.
    \item \textbf{FastText}: An efficient open-source library to learn text representations and text classifiers \cite{joulin2016bag}. 
    \item \textbf{SWEM}: Simple word embedding models \cite{shen2018baseline}, which employ simple pooling strategies operated over word embeddings.
    \item \textbf{LEAM}: Label-embedding attentive models \cite{wang2018joint}, which embed the words and labels in the same joint space for text classification.
    \item \textbf{Graph-CNN}: A graph CNN model that operates convolutions over word embedding similarity graphs \cite{defferrard2016convolutional}, where Chebyshev (C), Spline (S) and Fourier (F) filters are used.
    \item \textbf{Text-GCN} \cite{yao2018graph}: A graph convolution based model that puts both words and documents into one unified graph.
    \item \textbf{HAN} (Hierarchical Attention Networks)\cite{yang2016hierarchical} : A hierarchical LSTM model that encodes word and sentence vectors hierarchically.
    \item \textbf{Transformer} \cite{vaswani2017attention}: This model uses multi-head self attention to encode the words. 
    \footnote{ \url{https://github.com/harvardnlp/annotated-transformer}}
    \item \textbf{GraphSAGE} \cite{hamilton2017inductive}: As this model is originally designed for node classification problem, we add an attentive pooling layer on the top of the original model to represent the whole graph. 
    \item \textbf{S-LSTM}: A model that uses LSTM to exchange information between local and global nodes.
\end{itemize}

\subsection{Settings}
Words appear less than five times are replaced with \textit{UNK}. Sentences are truncated to 200. By default, the layer number is set to 6 for ReGNN, S-LSTM, Transformer for fair comparison. The default layer number for other graph-based models is 2. The hidden size is 300. We use Adam \cite{Kingma2015} optimizer to minimize the loss. The initial learning rate is 0.001, which is decayed by 0.5 every epoch. For datasets with small size, we set the batch size to 10. For datasets with big size, we set the batch size to 64. We train the models for 20 epochs and use the parameters with the best accuracy on the development set as the final parameters to test. For single label classification tasks, pre-trained 300-dimensional GloVe word embedding \cite{Pennington2014}\footnote{\url{http://nlp.stanford.edu/data/glove.6B.zip}} is used. For multi-label classification tasks, the word and label embeddings are randomly initialized. The maximum number of generated labels is set to 8.

\subsection{Experiment Results}

In Table \ref{tab:single result} we show the experiment results of single label text classification tasks. We can see that our proposed ReGNN surpasses all other baselines on four datasets and achieves the second best on Ohsumed. Furthermore, ReGNN performs more stable. The Text GCN model which achieves the highest accuracy on Ohsumed does not perform very well on datasets other than Ohsumed. 

One observation is that for single label text classification tasks, variations of graph based CNN (Graph-CNN, Text-GCN) models do not outperform traditional sequence based CNN. This phenomenon shows that the existing ways of neglecting the sequential information when modeling text in the form of graphs are problematic. Therefore, we combine the direct neighbors and word co-occurrence based neighbors together in the text graph. The Graph Convolutional Networks (GCN) suffer from over-smoothing problem \cite{li2018deeper}. When increasing the layer number, the representations of different nodes tend to become similar. However, deeper network means better ability of abstraction. The over-smoothing problem limits the abstraction ability of GCN. Therefore, we propose to use LSTM to help decide how to update the hidden states of the nodes. 

In Table \ref{tab:multi result} we show the experiment results of multiple label text classification tasks. From the results we can see that our proposed ReGNN model outperforms all other baseline models for the F1 score. S-LSTM and HAN are all strong baselines that produce competitive results. However, their performance is not as stable as ReGNN across different datasets. In multiple label classification tasks, we can not apply Text GCN model as a baseline because the sizes of these datasets are much too big to be fed into the memory.

\subsection{Analysis}
\begin{figure}[t]
    \centering
    \hspace{-5mm}
    \subfigure{
    \begin{minipage}[t]{0.5\linewidth}
    \centering
    \includegraphics[width=1.65in]{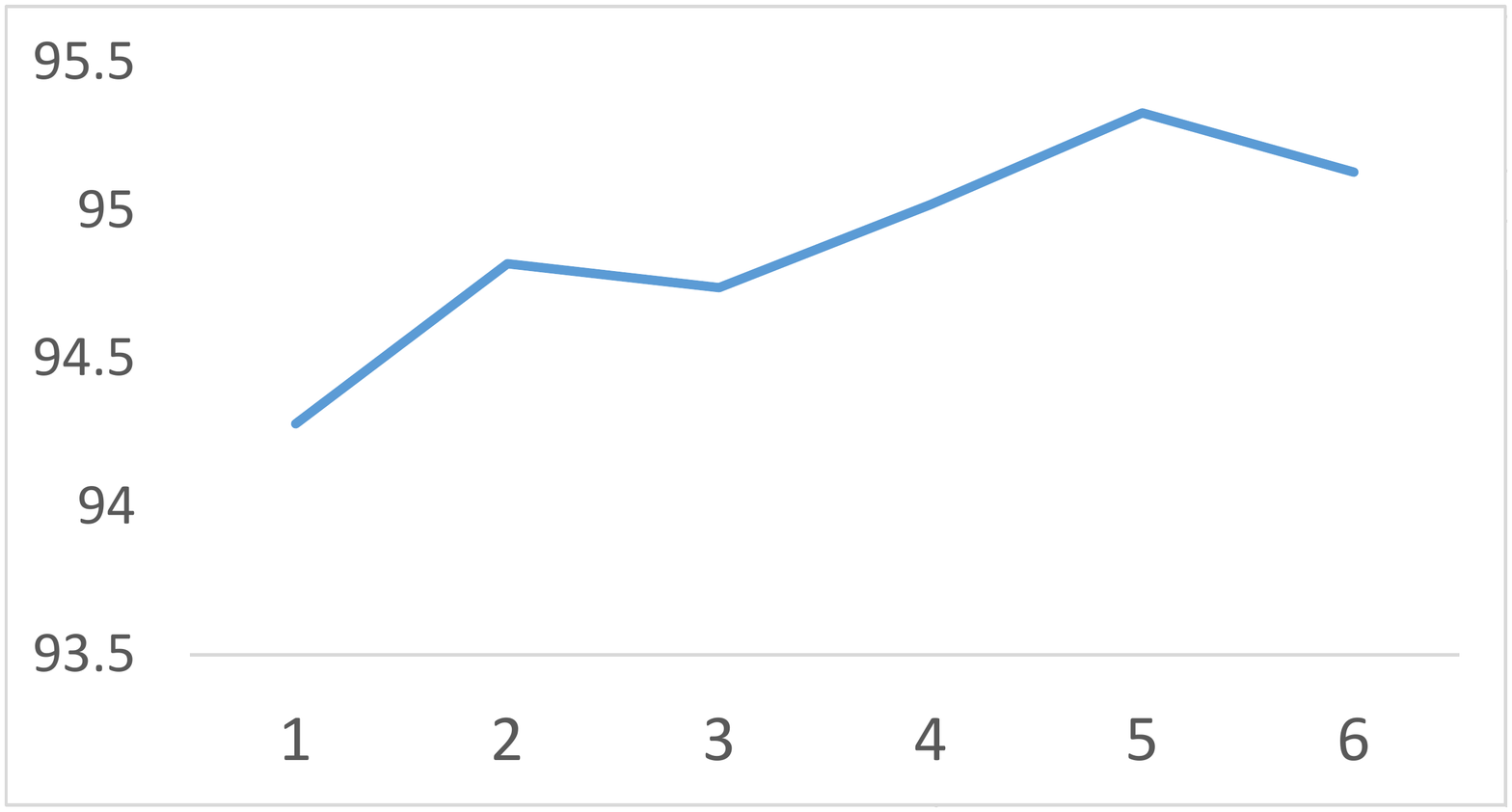}
    \end{minipage}%
    }%
    \subfigure{
    \begin{minipage}[t]{0.5\linewidth}
    \centering
    \includegraphics[width=1.65in]{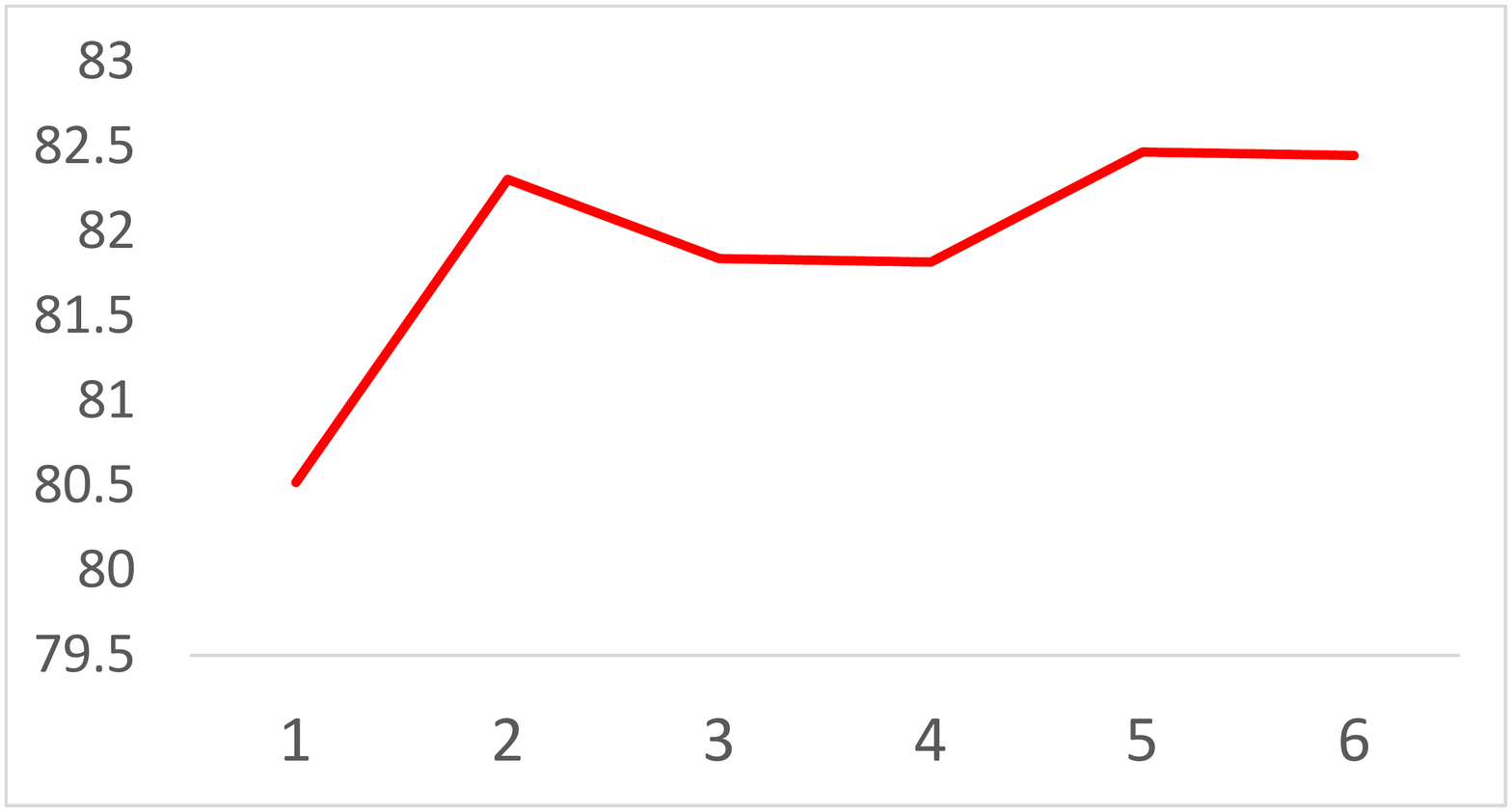}
    \end{minipage}%
    }%
    \caption{Accuracy of R52 ({\color{blue}left}) and Reuters21578 ({\color{red}right}) with different maximum neighbor numbers.}
    \label{fig:neighbor number}
\end{figure}

In Figure \ref{fig:neighbor number} we show the test accuracy with different maximum neighbor numbers in R52 (single label) and Reuters21578 (multiple label). To test the effect of word co-occurrence based graph, we only use the neighbors extracted with PMI scores. We can see that the general trend is that as the number of neighbors increases, the accuracy also increases until the number of neighbors achieves 5. This is expected because with more neighbors, the node can get access to more information. 
However, since the connection of neighbors is calculated based on the PMI score, the number of words with positive PMI score in the sentence (sliding window) is limited. Therefore, the text graph would stay the same and the performance would become steady.

\begin{table}[t]
    \centering
    \begin{tabular}{l|cc} \hline
     Model & R52 & Reuters21578 \\ \hline
     w/o LSTM & 84.74 & 43.82 \\
     w/o Attention & 94.39 & 81.31 \\
     w/o Global node & 93.85 & 76.81\\ \hline
     Proposal & 95.29 & 82.01 \\
     \hline
    \end{tabular}
    \caption{Ablation study on R52 and Reuters21578. It is clear that LSTM plays an important role by alleviating over-smoothing problem, especially in multi-label classification, which is more prone to over-smoothing.}
    \label{tab:ablation_study}
\end{table}
\subsubsection{Abalation Study}
In Table \ref{tab:ablation_study} we show the ablation study results by removing LSTM (Updation), attention (Aggregation) and graph-level node respectively. From the results we can see that removing any of the three parts of our proposed model would lead to a decline in accuracy. Among the three parts, the accuracy of the model without LSTM decreases most significantly. We assume that this is because that the over-smoothing problem becomes very severe with relatively big layer number. Furthermore, compared with multiple label classification, smoothing is more acceptable for single label classification. Because if the representation of all the nodes grow to be related to the correct label, it will not hurt the performance. However, over-smoothing is more harmful for multi-label problem, because multi-label requires the representation of different parts related to different labels to be distinguishable.

\subsubsection{Effectiveness on Over-smoothing Problem}
In this subsection, we give further analyzation on the effective of LSTM component on the  over-smoothing problem. We choose three graph-based models, the proposed ReGNN, ReGNN without LSTM component and GraphSAGE. 

\begin{figure}
    \centering
    \includegraphics[width=5cm]{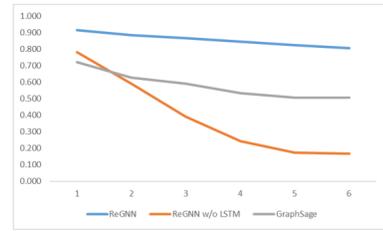}
    \caption{Effect of LSTM on \textbf{over-smoothing} problem. We test with three models, ReGNN, ReGNN without LSTM and GraphSAGE on Reuters21578. We show the average cosine distance among nodes as the layer number increases. This shows traditional GNN models without LSTM suffer from severe over-smoothing problem.}
    \label{fig:over-smoothing}
\end{figure}

In Figure \ref{fig:over-smoothing}, we show the effect of LSTM on the over-smoothing problem with the three graph-based model on Reuters21578 with different number of layers. The distance is measured by the average cosine distance between one node and all the others. The smaller the distance is, the more similar the representation among nodes are. From the figure we can see that as the layer number increases, the average distance of the two models without LSTM module decreases severely, which are all much lower than the distance of ReGNN. Appropriate amount of decline in distance is not hazard, which is the consequence brought by information propagation. However, the distance of models without LSTM decreases too much, which is described as over-smoothing. This phenomenon testifies our assumption that the GNN models without LSTM suffer from the over-smoothing problem and applying LSTM between layers can effectively alleviate the over-smoothing problem.

\begin{table}[t]
    \centering
    \begin{tabular}{p{7cm}}\hline
    \small
    \it the french ship [OOV] wallis , dwt , [OOV] at the port of [OOV] in victoria today to load tonnes of urgently needed \fancyhl{70}{wheat} for fiji after australian port unions partly lifted a trade embargo , shipping sources said . the wheat is expected to be loaded tomorrow , an australian \fancyhl{30}{wheat} board spokesman said . reuter \\ \hline
    \end{tabular}
    \caption{Attention heat map of the decoding process in Reuters21578. The gold label at the time is ``\textit{grain}''.}
    \label{tab:heat map}
\end{table}

In Table \ref{tab:heat map} we show an example of the attention heat map during the decoding process of Reuters21578. The gold label at the time is ``grain''. We can see that the model successfully pays attention to the word ``wheat'', which is highly related to the gold label ``grain''. The attention scores on other nodes are close to zero. If a model suffers from the over-smoothing problem, the hidden states of all nodes would result in very similar vectors, which means the attention scores would be averaged across all the nodes. This observation further testifies that our model does not suffer from the over-smoothing problem that many graph neural networks do \cite{zhou2018graph}. 

\section{Conclusion}
In this paper, we propose a recursive graphical neural network model to represent the text graph with dense vectors. The text graph is constructed via word co-occurrence information. We propose to use the LSTM component to dynamically decide which part of the neighbor information aggregated by attention mechanism should be involved for the updation of the hidden states. Experiment results on both single and multiple label text classification testify the effectiveness of our proposed model. Furthermore, we give detailed analysis to show that using LSTM between layers of GNN can effectively alleviate the over-smoothing problem, which is faced by many graph based models. It should be noted that our model is not limited to the classification task. In the future, we would like to apply our model to other tasks, such as generation tasks.

\bibliography{main}
\bibliographystyle{aaaibst}

\end{document}